\documentclass[10pt,twocolumn,letterpaper]{article}

\usepackage{cvpr}
\usepackage{times}
\usepackage{epsfig}
\usepackage{graphicx}
\usepackage{amsmath}
\usepackage{amssymb}

\usepackage{mathtools}
\usepackage{bm}
\usepackage{soul}
\usepackage{array,multirow,graphicx}
\usepackage{floatrow}
\newfloatcommand{capbtabbox}{table}[][\FBwidth]
\usepackage{wrapfig,lipsum,booktabs}
\usepackage[table,x11names,dvipsnames,table]{xcolor}

\usepackage[linesnumbered,ruled,vlined]{algorithm2e}
\usepackage{algorithmic}
\usepackage{xcolor}

\SetCommentSty{mycommfont}
\SetKwInput{Require}{Require}
\SetKwInput{Ensure}{Ensure}

\usepackage{siunitx}
\usepackage[export]{adjustbox} %
\usepackage{amssymb}%
\usepackage{pifont}%
\newcommand{\cmark}{\ding{51}}
\newcommand{\xmark}{\ding{55}}

\usepackage[pagebackref=true,breaklinks=true,letterpaper=true,colorlinks,bookmarks=false]{hyperref}
\cvprfinalcopy %

\begin{document}

\title{Semi-supervised Learning via Conditional Rotation Angle Estimation}

\author{
Hai-Ming Xu, ~ 
Lingqiao Liu, ~ 
Dong Gong \\
The University of Adelaide, Australia\\
{\tt\small \{hai-ming.xu;lingqiao.liu\}@adelaide.edu.au;edgong01@gmail.com}}

\maketitle
\begin{abstract}
   Self-supervised learning (SlfSL), aiming at learning feature representations through ingeniously designed pretext tasks without human annotation, has achieved compelling progress in the past few years. Very recently, SlfSL has also been identified as a promising solution for semi-supervised learning (SemSL) since it offers a new paradigm to utilize unlabeled data. This work further explores this direction by proposing to couple SlfSL with SemSL. Our insight is that the prediction target in SemSL can be modeled as the latent factor in the predictor for the SlfSL target. Marginalizing over the latent factor naturally derives a new formulation which marries the prediction targets of these two learning processes. By implementing this idea through a simple-but-effective SlfSL approach -- rotation angle prediction, we create a new SemSL approach called Conditional Rotation Angle Estimation (CRAE). Specifically, CRAE is featured by adopting a module which predicts the image rotation angle \textbf{conditioned on the candidate image class}. Through experimental evaluation, we show that CRAE achieves superior performance over the other existing ways of combining SlfSL and SemSL. To further boost CRAE, 
   we propose two extensions to strengthen the coupling between SemSL target and SlfSL target in basic CRAE. We show that this leads to an improved CRAE method which can achieve the state-of-the-art SemSL performance.
\end{abstract}

\section{Introduction}
The recent success of deep learning is largely attributed to the availability of a large amount of labeled data. However, acquiring high-quality labels can be very expensive and time-consuming. Thus methods that can leverage easily accessible unlabeled data become extremely attractive. Semi-supervised learning (SemSL) and self-supervised learning (SlfSL) are two learning paradigms that can effectively utilize massive unlabeled data to bring improvement to predictive models. 

SemSL assumes that a small portion of training data is provided with annotations and the research question is how to use the unlabeled training data to generate additional supervision signals for building a better predictive model. In the past few years, various SemSL approaches have been developed in the context of deep learning. The current state-of-the-art methods, e.g. MixMatch~\cite{berthelot2019mixmatch}, unsupervised data augmentation~\cite{li2018undeepvo}, converge to the strategy of combining multiple SemSL techniques, e.g. $\Pi$-Model~\cite{aila2016pimodel}, Mean Teacher~\cite{tarvainen2017mean}, mixup~\cite{zhang2017mixup}, which have been proved successful in the past literature. 

SlfSL aims for a more ambitious goal of learning representation without any human annotation. The key assumption in SlfSL is that a properly designed pretext predictive task which can be effortlessly derived from data itself can provide sufficient supervision to train a good feature representation. In the standard setting, the feature learning process is unaware of the downstream tasks, and it is expected that the learned feature can benefit various recognition tasks. SlfSL also offers a new possibility for SemSL since it suggests a new paradigm of using unlabeled data, i.e., use them for feature training. Recent work~\cite{s4l2019zhai} has shown great potential in this direction. 

This work further advances this direction by proposing to couple SlfSL with SemSL. The key idea is that the prediction target in SemSL can serve as a latent factor in the course of predicting the pretext target in a SlfSL approach. The connection between the predictive targets of those two learning processes can be established through marginalization over the latent factor, which also implies a new method of SemSL. Specifically, we implement this idea by extending the rotation angle estimation \cite{gidaris2018unsupervised} -- a recently proposed SlfSL approach for image recognition, and the key component of our method is a module that estimates the rotation angle conditioned on the class of the input image. Therefore, we call our method Conditional Rotation Angle Estimation (CRAE). To further promote the mutual dependency of the prediction task of SemSL and SlfSL, we propose two extensions: one sharpens the estimated posterior probability of image classes on unlabeled data and encourages the conditional rotation estimator being more specialized towards its conditioned class; another modifies the rotation estimation task to increase the difficulty of guessing the rotation angle without using class information, and thus make overall loss minimization depend more on correct estimation of the image class. 

Through experiments, we show that the proposed CRAE achieves significantly better performance than the other SlfSL-based SemSL approaches, and the extended CRAE  is on par with the state-of-the-art SemSL methods. In summary, the main contributions of this paper are as follows:
\begin{itemize}
    \item We propose a new SemSL idea by seamlessly coupling SlfSL and SemSL. Implementing this idea with a SlfSL approach, we create a new SemSL approach (CRAE) that can achieve superior performance than other SlfSL-based SemSL methods.
    \item We further make two extensions over the basic CRAE to boost its performance. The resulted new method achieves the state-of-the-art performance of SemSL. 
\end{itemize}

\section{Related Work}
Our work is closely related to both SemSL and SlfSL.

\textbf{SemSL} is a long-standing research topic which aims to learn a predictor from a few labeled examples along with abundant of unlabeled ones. SemSL based on different principals are developed in the past decades, e.g., "transductive" models~\cite{gammerman1998learning,joachims2003transductive}, multi-view style approaches~\cite{blum1998combining,zhou2005tri} and generative model-based methods~\cite{kingma2014semi,springenberg2015unsupervised}, etc. Recently, the consistency regularization based methods have become quite influential due to their promising performance in the context of deep learning. Specifically, $\Pi$-Model~\cite{aila2016pimodel} requires model's predictions to be invariant when various perturbations are added to the input data. Mean Teacher~\cite{tarvainen2017mean} enforces a student model producing similar output as a teacher model whose weights are calculated through the moving average over the weight of student model. Virtual Adversarial Training~\cite{miyato2018virtual} encourages the predictions for input data and its adversarially perturbed version to be consistent. More recently, mixup~\cite{zhang2017mixup,verma2019interpolation} has emerged as a powerful SemSL regularization method which requires the output of mixed data to be close to the output mixing of original images. In order to achieve good performance, most state-of-the-art approaches adopt the strategy of combining several existing techniques together. For example, Interpolation Consistency Training~\cite{verma2019interpolation} incorporates Mean Teacher into the mixup regularization, MixMatch~\cite{berthelot2019mixmatch} adopts a technique that uses fused predictions as pseudo prediction target as well as the mixup regularization. Unsupervised data augmentation~\cite{li2018undeepvo} upgrades $\Pi$-Model with advanced data augmentation methods.

\textbf{SlfSL} is another powerful paradigm which learns feature representations through training on pretext tasks whose labels are not human annotated~\cite{kolesnikov2019revisiting}. Various pretext tasks are designed in different approaches. For example, image inpainting~\cite{pathak2016context} trains model to reproduce an arbitrary masked region of the input image.
Image colorization~\cite{zhang2016colorful} encourages model to perform colorization of an input grayscale image. Rotation angle prediction~\cite{gidaris2018unsupervised} forces model to recognize the angle of a rotated input image. After training with the pretext task defined in a SlfSL method, the network is used as a pretrained model and can be fine-tuned for a downstream task on task-specific data. Generally speaking, it is still challenging for SlfSL method to achieve competitive performance to fully-supervised approaches. However, SlfSL provides many new insights into the use of unlabeled data and may have a profound impact to other learning paradigms, such as semi-supervised learning.

\textbf{SlfSL based SemSL} is an emerging approach which incorporates SlfSL into SemSL. The most straightforward approach is to first perform SlfSL on all available data and then fine-tune the learned model on labeled samples. $S^4L$~\cite{s4l2019zhai} is a newly proposed method which jointly train the downstream task and pretext task in a multi-task fashion without breaking them into stages. In this paper, we further advance this direction through proposing a novel architecture which explicitly links these two tasks together and ensure that solving one task is beneficial to the other.

\begin{figure}[t]
\centering
\includegraphics[width=\textwidth]{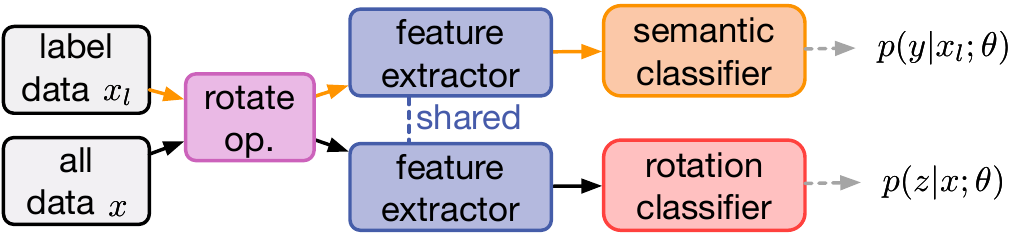}
\caption{Structure of $S^4L$ method.}
\label{fig:s4l}
\end{figure}
\section{Our approach}
\subsection{Coupling SemSL with SlfSL}\label{sect:framework}
\begin{figure*}[t]
\centering
\includegraphics[width=0.9\textwidth]{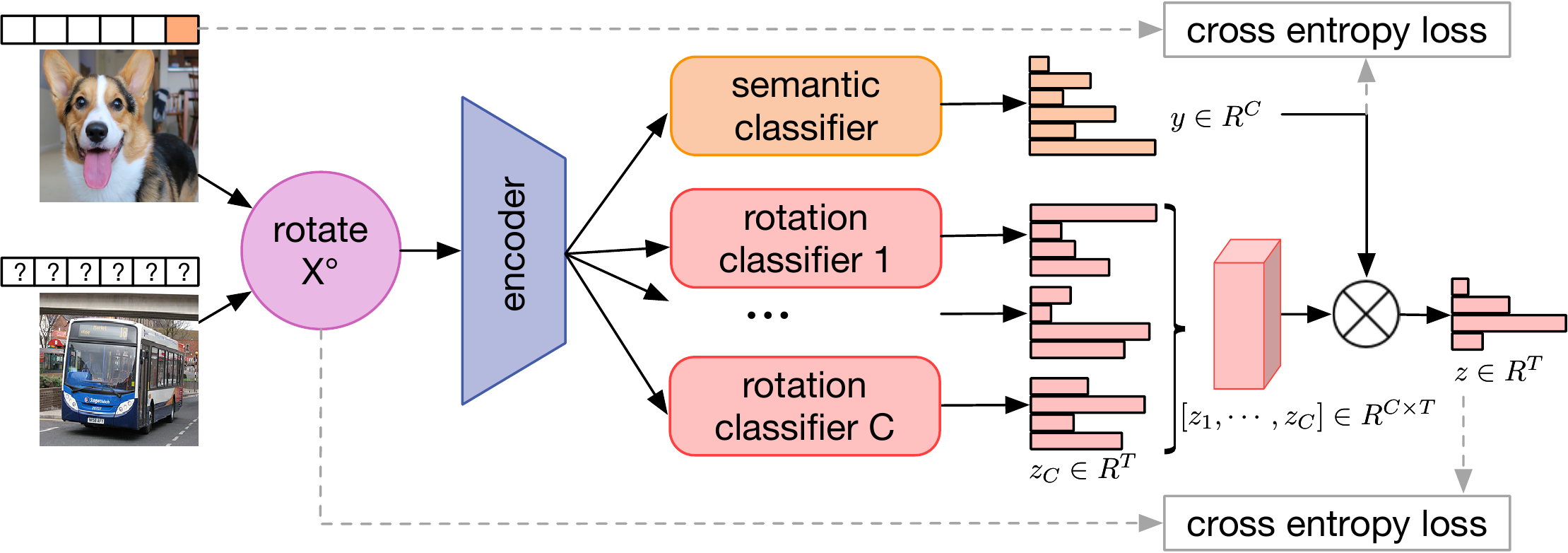}
\caption{Illustration of our proposed CARE method. CRAE creates multiple conditional rotation estimation branches, one for each class. The semantic classifier acts as a switch to softly select (through weighted summation) the estimation of $z$ from each rotation classifier.}

\label{fig:conditionalrot}
\end{figure*}
In SemSL, we are given a set of training samples $\{x_1, x_2, \cdots, x_n\}\in~X$ with only a few of them $X_l=\{x_1, x_2, \cdots, x_l\}\in~X$ annotated with labels $\{y_1, y_2, \cdots, y_l\}\in~Y_l$ (usually $l\ll n$ and $y$ is considered as class label here). The goal of a SemSL algorithm is to learn a better posterior probability estimator over $y$ from both labeled and unlabeled training samples, i.e., learning $p(y|x,\theta)$ with $\theta$ as model parameters,  SlfSL aims to learn feature representations via a pretext task. The task usually defines a target $z$, which can be derived from the training data itself, e.g., rotation angle of the input image. Once $z$ is defined, SlfSL is equivalent to training a predictor to model $p(z|x; \theta)$. There are two existing schemes to leverage SlfSL for SemSL. The first is to use SlfSL to learn the feature from the whole training set and then fine-tune the network on the labeled part. The other is jointly optimizing the tasks of predicting $y$ and $z$, as in the recently proposed  $S^4L$ method, shown in Figure \ref{fig:s4l}. $S^4L$ constructs a network with two branches and a shared feature extractor. One branch for modeling $p(y|x; \theta)$ and another for $p(z|x; \theta)$. However, in both methods the prediction $p(z|x; \theta)$ of the pretext target $z$ is implicitly related to the task of predicting $y$.

In this work, we are interested in an explicit connection between the prediction target $z$ and $y$. To establish this connection, we treat $y$ as the latent factor in $p(z|x; \theta)$ and factorize $p(z|x; \theta)$ through marginalization:
\begin{equation}\label{conditionalRot}
p(z|x; \theta) = \sum_y p(z, y|x; \theta) = \sum_y p(z|x, y; \theta)p(y|x; \theta).
\end{equation} Eq.~\ref{conditionalRot} suggests that the pretext target predictor $p(z|x; \theta)$ can be implemented as two parts: a model to estimate $p(y|x; \theta)$ and a model to estimate $z$ conditioned on both $x$ and $y$, i.e., $p(z|x, y; \theta)$. For the labeled samples, the ground-truth $y$ is observed and can be used for training $p(y|x; \theta)$. For unlabeled samples, the estimation from $p(y|x; \theta)$ and $p(z|x, y; \theta)$ will be combined together to make the final prediction about $z$. 
Consequently, optimizing the loss for $p(z|x; \theta)$ will also provide gradient to back-propagate through $p(y|x; \theta)$. This is in contrast to the case of $S^4L$, where the gradient generated from the unlabeled data will not flow through $p(y|x; \theta)$.

\subsection{CRAE: SemSL via conditional rotation angle estimation}\label{sect:CRAP}
In the following part, we describe our implementation of the above idea. Eq.~\ref{conditionalRot} may apply to a range of SlfSL approaches. In this work, we employ 
rotation angle estimation~\cite{gidaris2018unsupervised}-- a recently proposed SlfSL approach for image recognition. It randomly rotates the input image by one of the four possible rotation angles ($\{\ang{0}, \ang{90}, \ang{180}, \ang{270}\}$) and requires the network to give a correct prediction of the rotation angle. Despite being extremely simple, this method works surprisingly well in practice. The underlying logic is that to correctly predict the rotation angle, the network needs to recognize the canonical view of objects from each class and thus enforces the network to learn informative patterns of each image category.

There are two components in the Eq.~\ref{conditionalRot}: one is $p(y|x)$ and the other is $p(z|y,x)$ (hereafter, $\theta$ is omitted for simplicity). We implement both components as prediction heads attaching to the same feature extraction backbone. $p(y|x)$ is simply implemented as a classifier branch. For rotation angle estimation, the pretext target $z$ is the rotation angle of the input image. $p(z|x,y)$ is thus a rotation angle estimator conditioned on a candidate image class $y$. In this work, we realize $p(z|x, y)$ by allocating a rotation angle estimation branch for each candidate class. Since this method is featured by the conditional rotation angle estimation (CRAE), we call our method CRAE.
To simplify the notation, we denote each rotation angle estimation branch as $R_k(x) = p(z|x,y=k), ~k= 1\cdots C$ or simply $R_k$. Follow Eq.~\ref{conditionalRot}, the prediction from each branch is then aggregated with the aid of $p(y|x)$ for the final prediction of $z$, that is, $p(z|x) = \sum_{k=1}^C R_k(x) p(y=k|x)$. A more detailed schematic illustration of the CRAE method is shown in Figure~\ref{fig:conditionalrot}. Note that after training, all the rotation angle estimation branches are discarded. Thus our method will not introduce additional computational cost at the test stage.  

Formally, the training objective of CRAE is given as follows:
\begin{small}
\begin{equation}\label{obj_func}
L = \sum_{(x_l,y_l) \in \mathcal{L}}L_{ce}(p(y|x_l), y_l) + \eta \sum_{x \in \mathcal{U} \cup \mathcal{L}}L_{ce} (p(z|x), \hat{z} ), 
\end{equation}
\end{small}where $\mathcal{L}$ and $\mathcal{U}$ are labeled and unlabeled datasets respectively. $\hat{z}$ is the ground truth of rotation angle. $L_{ce}$ denotes the cross-entropy loss. Note that the estimation of $p(z|x)$ involves the term $p(y|x)$ as shown in Eq. \ref{conditionalRot}. For labeled data, we use ground-truth $y_l$, i.e., the one-hot vector to replace the estimated $p(y|x_l)$.

\subsection{Analyzing the training process of CRAE}\label{sect:analysis}
In Eq.~\ref{conditionalRot}, the benefit of our method was explained as being able to generate gradient to update $p(y|x)$ on unlabeled data. But why this ability will lead to better estimation of $p(y|x)$? The underlying mechanism is actually more involved. In the following, we analyze the training process of CRAE with more details. 

\begin{figure}[t]
\centering
\includegraphics[width=\textwidth]{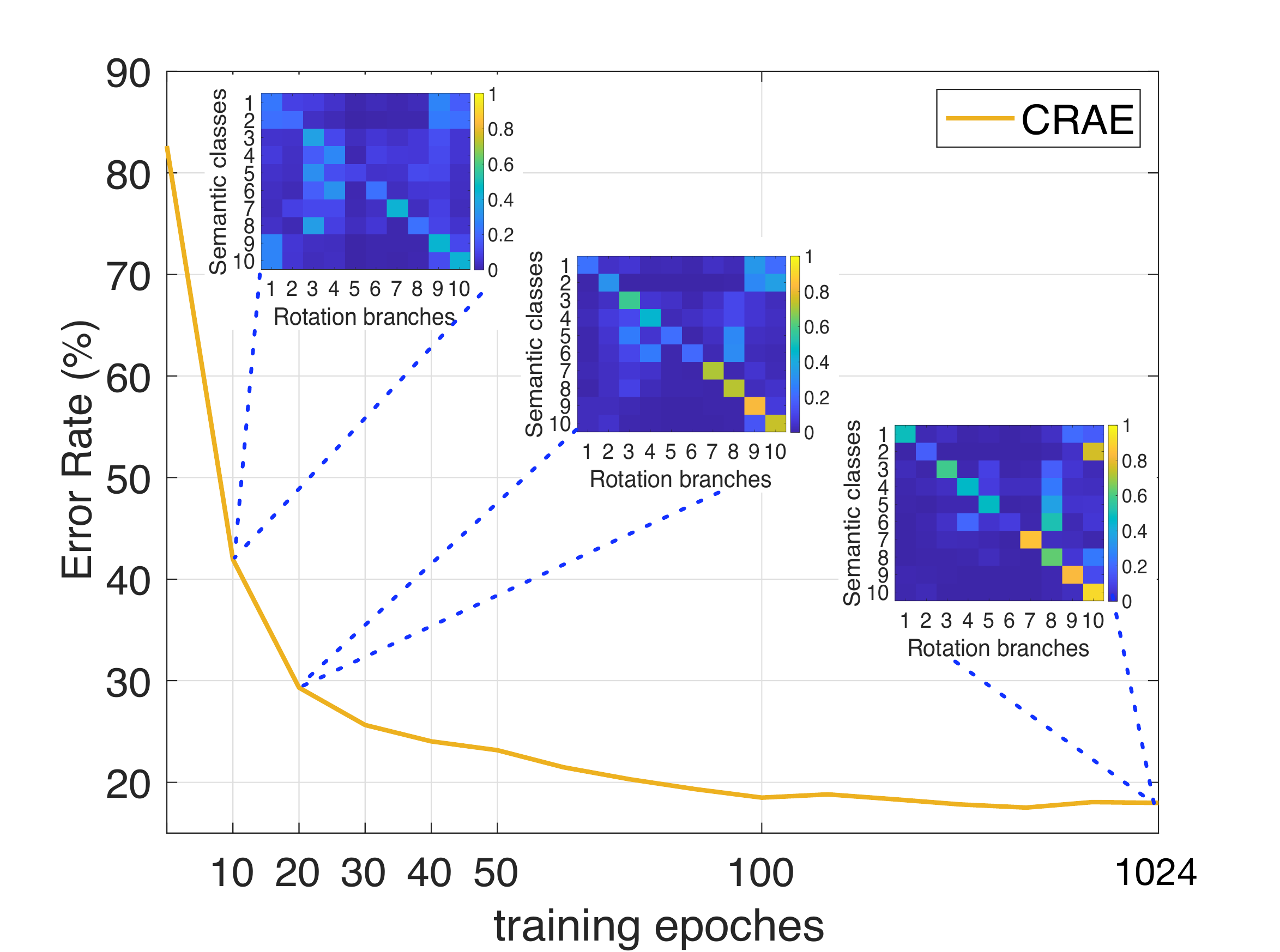}
\caption{Visualization for error rates (\%) of classifier and confusion matrix change during the training progress of CRAE on CIFAR-10 with 250 labels. If the confusion matrix is closer to a diagonal matrix, it indicates $R_k$ is more specialized to class $k$. See the main text for more details. Here, three confusion matrices, corresponding to 10, 20, 1024 epoch respectively, are measured.}

\label{fig:diversity}
\end{figure}

There are three key insights which are essential to understand CRAE: \textbf{(1) a good estimation of $p(y|x)$ will make each $R_k$ more specialized to the $k$-th class.} In Eq.~\ref{conditionalRot}, $p(y|x)$ can be viewed as a switch to (softly) select the relevant $R_k$. When $p(y|x)$ is close to the ground-truth distribution -- an one-hot vector (with the only ``1'' given to the correct class of $x$), only the $R_k$ corresponding to the correct class will be used for predicting $z$ and being updated. 
This will make $R_k$ be updated solely (or more adequately in reality) for images from the $k$-th class, and consequently $R_k$ tends to produce better estimation of $z$ if the class label of $x$ is $k$. \textbf{(2) on the other hand, a specialized $R_k$ will push $p(y|x)$ to give higher probability to the true class of $x$, and thus improve the estimation of $p(y|x)$.} This is because in order to optimize the overall estimation of $z$, the training process tends to select $R_k$ whose prediction is closest to the ground-truth rotation angle.
Follow the conclusion in (1) -- $R_k$ tends to give better estimation of $z$ if the true image class is $k$, the training process will push $p(y|x)$ to attain higher value for the true class.
\textbf{(3) the update of $R_k$ and the branch for estimating $p(y|x)$ will help learning better feature representations.} $R_k$ is essentially a rotation angle estimator, according to \cite{gidaris2018unsupervised}, optimizing the rotation angle estimation will induce a better feature representation. $p(y|x)$ is directly related to predicting target of interest. The update of $p(y|x)$ will simultaneously improve the classifier branch and its attached feature backbone. In some sense, updating feature representation through $p(y|x)$ is ideal since ultimately the feature will be used for predicting $y$.
In summary, we can see that the update of $p(y|x)$ and $R_k$ is mutually dependent and reinforce to each other.

In practice, the estimation of $p(y|x)$ is far from perfect at the beginning of training due to the limited number of labeled samples. At this stage, $R_k$ is less specialized towards their corresponding class. To validate this, we plot the prediction error rates of the semantic classifier branch on unlabeled data and the confusion matrix in Figure \ref{fig:diversity}, where the $(i,j)$ entry of the confusion matrix indicates the chance that $R_j$ gives the best estimation for $z$ given the true class being $i$. It indicates $R_k$ is more specialized to $k$ if it is more close to a diagonal matrix. As seen, at the first few epochs, the error rates of $p(y|x)$ is high and $R_k$ is less specialized.

However, since $R_k$ also has the effect of learning features, the shared feature extractor will be improved through the course of training. With a better feature representation, the estimation of $p(y|x)$ will also be improved. The improved $p(y|x)$ branch will then follow the analysis above to make $R_k$ specialized for $k$ and this will in turn reinforce the estimation of $p(y|x)$. 
As seen from Figure~\ref{fig:diversity}, the error rates of $p(y|x)$ decreases and the confusion matrix becomes closer to diagonal matrix with more training epochs.

\subsection{Extensions to CRAE}\label{extension}
In the following, we propose two extensions to further improve the training process of CRAE from the perspectives of $p(y|x)$ and $p(z|x,y)$, respectively.

\noindent \textbf{Extension 1: Encouraging a sharper $p(y|x)$}

Based on analysis in Section \ref{sect:analysis}, we can see that $p(y|x)$ acts as a (soft) selector for $R_k$. A sharper $p(y|x)$ estimation, that is, an estimation of $p(y|x)$ closer to an one hot vector, will encourage $R_k$ to be updated mainly from its corresponding class, and thus make it more specialized to $k$. In this section, we create an additional loss to encourage $p(y|x)$ to be a sharper distribution. Specifically, we rotate each image in four angles within one batch (the predictions of the rotated image are the byproduct of CRAE) and obtain the average $\bar{p}$ of the predicted distributions across these four rotated samples. Then we perform a sharpening operation over $\bar{p}$ to create a sharpened target $\hat{p}$ for $p(y|x)$: \begin{align}\label{sharpening}
\hat{p_i} = \frac{(\bar{p}_i)^{\frac{1}{T}}}{\sum_{j=1}^C (\bar{p}_j)^{\frac{1}{T}}},
\end{align}where $C$ is the number of classes and $T\in(0, 1]$ is a temperature hyper-parameter. Then we use the cross entropy between $\hat{p}$ and $p(y|x)$ as an additional loss to encourage a sharper $p(y|x)$. Note that since the $\hat{p}$ is obtained by averaging estimations from four augmented input images, it is expected to be more accurate than $p(y|x)$. Thus this loss also has the effect of encouraging a more reliable estimation of $p(y|x)$. Empirically, we find it works better than other sparse regularizers such as entropy minimization loss. 
\noindent \textbf{Extension 2: Making $p(z|x)$ depend more on $p(y|x)$ }

CRAE provides indirect supervision for $p(y|x)$ based on the condition that the correct estimation of $p(y|x)$ will improve the overall estimation of $z$. Thus if the optimization of the overall estimation $p(z|x)$ relies more on the correct estimation of $p(y|x)$, then CRAE is more likely to provide stronger supervision for $p(y|x)$. 

To achieve this goal, we propose another extension by modifying the conditional rotation estimation task. Specifically, we require the rotation prediction branch to predict rotation angle for a mixed version of the rotated image, that is, we randomly mix the input image $x_i$ with another randomly sampled rotated image $x_j$ via $x_{\text{mix}} = \alpha x_i + (1-\alpha) x_j$, with $\alpha$ sampled from $[0.5, 1]$. Meanwhile, the class prediction $p(y|x_i)$ is calculated from the unmixed version of the input $x_i$. In such a design, the network needs to recognize the rotation angle of the target object with the noisy distraction from another image. The purpose of introducing this modified task is to make the SlfSL task more challenging and more dependent on the correct prediction from $p(y|x)$. 

To see this point, let's consider the following example. Letter `A' is rotated with \ang{270} and is mixed with letter `B' with rotation \ang{90}. Directly predicting the rotation angle for this mixed image encounters an ambiguity: whose rotation angle, A's or B's, is the right answer? In other words, the network cannot know which image class is the class-of-interest. This ambiguity can only be resolved from the output of $p(y|x)$ since its input is unmixed target image. Therefore, this improved rotation prediction task relies more on the correct prediction from the semantic classifier and training through CRAE is expected to give stronger supervision signal to $p(y|x)$. Note that although this scheme also uses mix operation, \textbf{it is completely different from mixup}~\cite{zhang2017mixup}. The latter constructs a loss to require the output of the mixed image to be mixed version of the outputs of original images. This loss is not applied in our method. 

In summary, the training objective for the extended CRAE is as follows
\begin{align}\label{obj_func}
L &= \sum_{(x_l,y_l) \in \mathcal{L}}L_{ce}(p(y|x_l), y_l) + \eta_1 \sum_{x \in \mathcal{U} \cup \mathcal{L}}L_{ce} (p(z|x_{\text{mix}}), \hat{z} ) \nonumber \\
& + \eta_2 \sum_{x_u \in \mathcal{U}} L_{\text{ce}}(p(y|x_u), \hat{p})
\end{align}
where $x_{\text{mix}}$ indicates the mixed input image and $\hat{p}$ in the last term is the sharpened target in Eq~\ref{sharpening}.

\subsection{Other Implementation details}\label{sec:obstacles}
One potential obstacle of our model is that the quantity of parameters in the rotation branches would increase significantly with a large $C$. To tackle this, we propose to perform dimension reduction for the features feeding into the rotation predictors. Results in Figure~\ref{fig:imagenet_dim_red} show that this scheme is effective as our performance will not drop even when the dimension is reduced from 2048 to 16. Also, we find for basic CRAE, it is beneficial to add an additional semantic classifier branch to the feature extractor, and learn it via the loss incurred from the labeled data only. At test time, this classifier will be directly used for testing. This treatment is similar to $S^4L$ and can bring a slight performance gain for basic CRAE. We postulate the reason is that the indirect supervision generated for the original $p(y|x)$ branch from CRAE training can be noisy comparing with supervision directly generated from the ground-truth $y$.
It is better to use such a branch just for feature learning since feature learning is more tolerant to noisy supervision.
More implementation details can be found in the supplementary material.

\section{Experiments}
In this section, we conduct experiments to evaluate the proposed CRAE method
. The purpose of our experiments is threefolds: (1) to validate if CRAE is better than other SlfSL-based SemSL algorithms. (2) to compare \textbf{CRAE} and extended CRAE (denoted as \textbf{CRAE+} hereafter) against the state-of-the-art SemSL methods. (3) to understand the contribution of various components in CRAE and CRAE+.

\subsection{Experimental details}
To make a fair comparison to recent works, different experimental protocols are adopted for different datasets. Specifically, for CIFAR-10 and CIFAR-100~\cite{krizhevsky2009learning} and SVHN~\cite{netzer2011reading}, we directly follow the settings in~\cite{berthelot2019mixmatch}. For ILSVRC-2012~\cite{russakovsky2015imagenet}, our settings are identical to~\cite{s4l2019zhai} except for data pre-processing operations for which we only use the inception crop augmentation and horizontal mirroring. We ensure that all the baselines are compared under the same setting. 
Followed the standard settings of SemSL, the performance with different amount of labeled samples are tested. For CIFAR-10 and SVHN, sample size of labeled images is ranged in five levels: \{250, 500, 1000, 2000, 4000\}. For CIFAR-100, 10000 labeled data is used for training. For ILSVRC-2012, 10\% and 1\% of images are labeled among the whole dataset. 
More experimental details can be found in the supplementary material.

\subsection{Comparing with SlfSL-Based SemSL methods}
Firstly, we compare CRAE to other SlfSL-based SemSL algorithms on five datasets: CIFAR-10, CIFAR-100, SVHN, SVHN+Extra and ILSVRC-2012.

Two SlfSL-based SemSL baseline approaches are considered: 1) \textbf{Fine-tune}: taking the model pre-trained on the pretext task as an initialization and fine-tuning with a set of labeled data. We term this method Fine-tune in the following sections. 2) $\bm{S^4L}$: $S^4L$ method proposed in~\cite{s4l2019zhai}. Note that our extension 1 uses additional training loss other than the SlfSL loss on the labeled data, and extension 2 uses a modified SlfSL task. Those factors are not considered in other SlfSL-based SemSL baselines. To make a fair comparison, we only use our basic CRAE model in the comparison in this subsection. 
Also, as a reference, we report the performance obtained by only using the labeled part of the dataset for training, denoting as \textbf{Labeled-only}.
The experimental results are as follows:

\noindent \textbf{CIFAR-10} The results are presented in Table~\ref{tab:cifar10}. We find that the ``Fine-tune'' strategy leads to a mix amount of improvement over the ``Labeled-only'' case. It is observed that a large improvement can be obtained when the amount of labeled samples is ranged from 500 to 2000 but not on 250 and 4000's settings. It might be because on the one hand too few labeled samples are not sufficient to perform an effective fine-tuning while on the other hand the significant improvement diminishes after the sample size increase. 
In comparison, $S^4L$ achieves much better accuracy for the case of using few samples. This is largely benefited from its down-stream-task awareness design. Our CRAE method achieves significantly better performance than those two ways of incorporating SlfSL for SemSL and always halves the test error of $S^4L$ in most cases.
\begin{table}[tbp]%
\centering
\caption{Error rates (\%) for SlfSL based SemSL on CIFAR-10.}
\label{tab:cifar10}
\begin{tabular}{l r r r r r }
\hline
\hline
\# Labels & \multicolumn{1}{c}{250} & \multicolumn{1}{c}{500} & \multicolumn{1}{c}{1000} & \multicolumn{1}{c}{2000} & \multicolumn{1}{c}{4000} \\
\hline
Labeled-only &  56.76 &	47.24 & 36.09 & 29.90 & 19.65 \\

Fine-tune & 53.66 & 35.18 & 28.17 & 22.00 & 17.39 \\

$S^4L$ & 32.66 & 28.23 & 22.55 & 18.91 & 15.71 \\

CRAE & \textbf{17.07} & \textbf{15.57} & \textbf{12.90} & \textbf{10.72} & \textbf{9.07} \\
\hline
\hline
\end{tabular}
\end{table}
\begin{table}[tbp]\small%
\centering
\caption{Error rates (\%) for SlfSL based SemSL on SVHN and SVHN+Extra.}
\label{tab:svhn}
\begin{tabular}{ l l r r r r r }
\hline
\hline
\multicolumn{2}{c}{\# Labels} & \multicolumn{1}{c}{250} & \multicolumn{1}{c}{500} & \multicolumn{1}{c}{1000} & \multicolumn{1}{c}{2000} & \multicolumn{1}{c}{4000} \\
\hline
\parbox[t]{2mm}{\multirow{4}{*}{\rotatebox[origin=c]{90}{SVHN}}} & Labeled-only &  23.27 & 17.85 & 12.84 & 9.65 & 7.26 \\

& Fine-tune & 16.21 & 11.54 & 8.31 & 7.30 & 5.61 \\

& $S^4L$ & 10.48 & 8.32 & 7.21 & 6.15 & 5.74 \\

& CRAE & \textbf{9.23} & \textbf{7.61} & \textbf{5.76} & \textbf{4.93} & \textbf{4.74} \\
\hline
\hline
\parbox[t]{2mm}{\multirow{3}{*}{\rotatebox[origin=c]{90}{+Extra}}} & Fine-tune & 13.11 & 9.52 & 7.74 & 6.17 & 5.57 \\

& $S^4L$ & 8.47 & 6.58 & 5.44 & 4.30 & 3.87 \\

& CRAE & \textbf{8.24} & \textbf{5.42} & \textbf{4.61} & \textbf{4.13} & \textbf{3.79} \\
\hline
\hline
\end{tabular}
\end{table}
\begin{table}[t]%
\centering
\caption{Error rates (\%) for SlfSL based SemSL on CIFAR-100.}
\label{tab:cifar100}
\begin{tabular}{ c c c c c}
\hline
\hline
Methods & Labeled-only & Fine-tune & $S^4L$ & CRAE \\
\hline
CIFAR-100 & 46.53 & 34.11 & 33.89 & \textbf{30.68} \\
\hline
\hline
\end{tabular}
\end{table}

\noindent \textbf{SVHN and SVHN+Extra} Table~\ref{tab:svhn} shows the results of each method. Apparently, both Fine-tune and $S^4L$ methods can achieve much better performance than the Labeled-only baseline especially for case with few labeled samples, i.e., only 250 or 500 labeled data. $S^4L$ outperforms Fine-tune method on most of the settings except for obtaining comparable results on 4000 labels setting. In comparison, the proposed CRAE still manages to produce superior accuracy than $S^4L$ on all those settings.
With more unlabeled samples from SVHN's Extra dataset, all methods gain a further improvement. Continuously, our method slightly excels both Fine-tune and $S^4L$ methods.

\noindent \textbf{CIFAR-100} As shown in Table~\ref{tab:cifar100}, it is obvious that all SlfSL-based SemSL methods can have better accuracy than that of Labeled-only and $S^4L$ leads to a marginal improvement over Fine-tune. Again, the proposed CRAE method performs better than those baselines. 

\begin{table}[t]%
\centering
\caption{ILSVRC-2012 accuracies (\%) for SlfSL based SemSL.}
\label{tab:imagenet}
\begin{tabular}{ l r r | r r }
\hline
\hline
\# Labels & \multicolumn{2}{c|}{10\%} & \multicolumn{2}{c}{1\%} \\
& Top1 & Top5 & Top1 & Top5 \\
\hline
Labeled-only &  - & 80.43 & - & 48.43 \\

Fine-tune & - & 78.53 & - &  45.11 \\

$S^4L$ & -	& 83.82	& -	& 53.37 \\
\hline
\hline
Labeled-only & 59.16 & 83.07 & 41.26 & 69.07 \\
$S^4L$ & 63.84 & 86.28 & 46.90 & 74.16 \\
CRAE & \textbf{65.34} & \textbf{87.07} & \textbf{49.26} & \textbf{75.57} \\
\hline
\hline
\end{tabular}
\end{table}
\begin{figure}[t]
\centering
\includegraphics[width=0.8\textwidth]{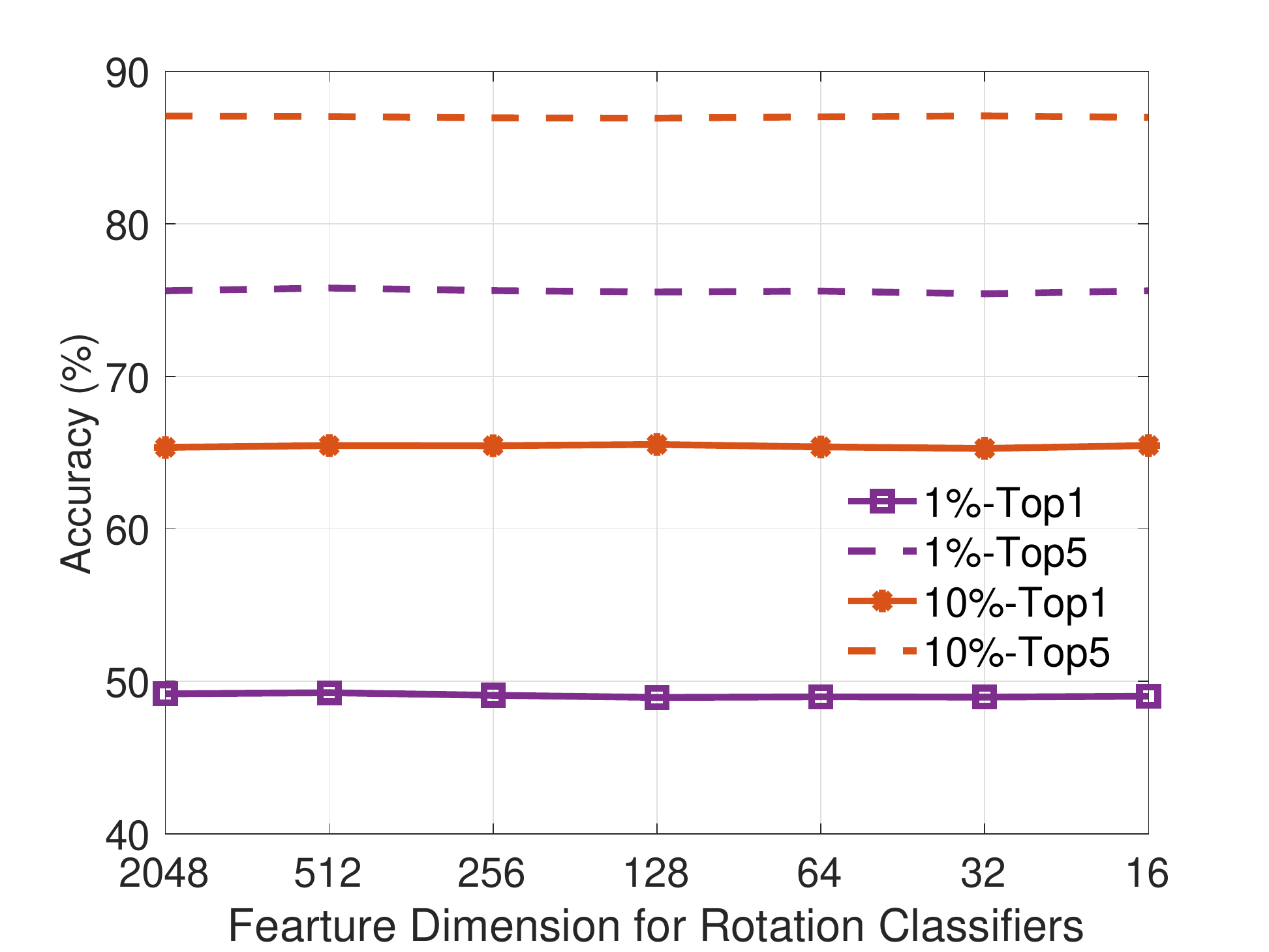}
\caption{ILSVRC-2012 accuracy (\%) changes with various dimension sizes of input feature channel for CRAE.}
\label{fig:imagenet_dim_red}
\end{figure}

\noindent \textbf{ILSVRC-2012} Table~\ref{tab:imagenet} presents the results of each method. The top block of Table~\ref{tab:imagenet} shows the reported results in the original $S^4L$ paper~\cite{s4l2019zhai} and we also re-implement $S^4L$ based on the code of~\cite{kolesnikov2019revisiting}. Due to the difference of data pre-processing, results in the upper block cannot be directly compared to those below. Again, we have observed that CRAE is consistently superior to $S^4L$ in all settings.

As mentioned in Section~\ref{sec:obstacles}, for saving the computational cost, we propose to reduce the dimensionality of features fed into the rotation angle predictors when there is a large number of classes. In Figure~\ref{fig:imagenet_dim_red}, we demonstrates the effect of this scheme. As seen, the test performance stays the same when the feature dimension is gradually reduced from 2048 to only 16 dimensions. This clearly validates the effectiveness of the proposed scheme.

\begin{figure}[t]
\centering
\includegraphics[width=\textwidth]{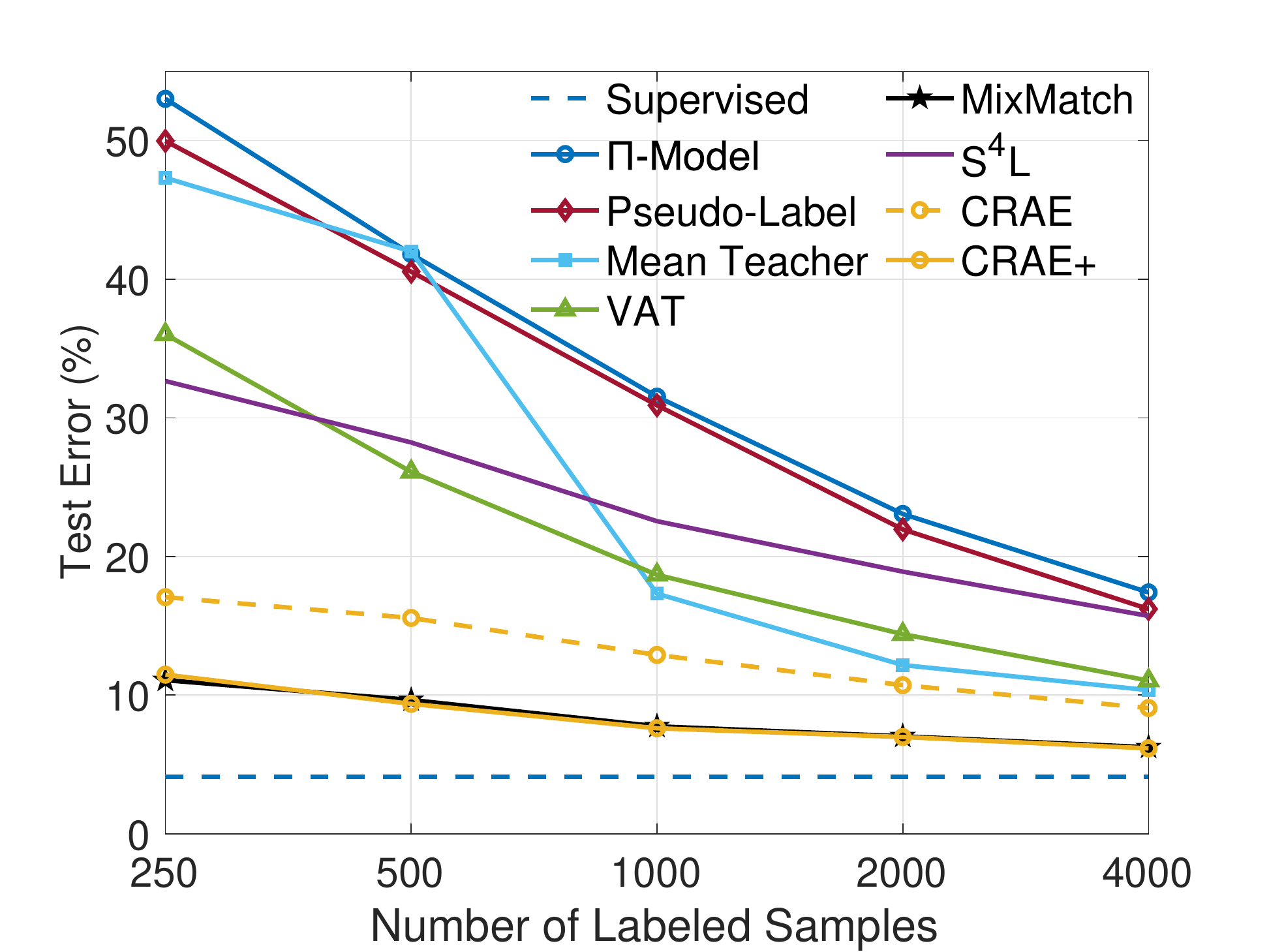}
\caption{CIFAR-10 error rates (\%) of CRAE+
and SOTA SemSL.}
\label{fig:cifar10}
\end{figure}

\begin{table}%
\centering
\caption{Error rates (\%) comparison of CRAE+
to SOTA SemSL methods on CIFAR-100.}
\label{tab:cifar100_ssl}%
\begin{tabular}{ c c c c c}
\hline
\hline
Methods & SWA & MixMatch & CRAE+ \\
\hline
CIFAR-100 & 28.8 & 25.88 & \textbf{25.69} \\
\hline
\hline
\end{tabular}
\end{table}

\subsection{Comparing with the state-of-the-art SemSL}\label{exp:compare_to_sota_ssl}
In the following section, we proceed to demonstrate the performance of CRAE+, that is, the extended CRAE method by incorporating the two extensions discussed in Section~\ref{extension}. We compare its performance against the current state-of-the-art methods in SemSL.
Similar to~\cite{berthelot2019mixmatch}, several SemSL baselines are considered: Pseudo-Label, $\Pi$-Model, Mean Teacher, Virtual Adversarial Training (VAT) and MixMatch\footnote{For CIFAR-100, we only compare CRAE+ against SWA~\cite{athiwaratkun2018there} and MixMatch for their achieving the best reported performance in literature.}. Since a fair and comprehensive comparison has been done in~\cite{berthelot2019mixmatch}, we strictly follow the same experimental setting and directly compare our proposed methods CRAE and CRAE+ to the numbers reported in~\cite{berthelot2019mixmatch}.

\begin{figure}[t]
\centering
\includegraphics[width=\textwidth]{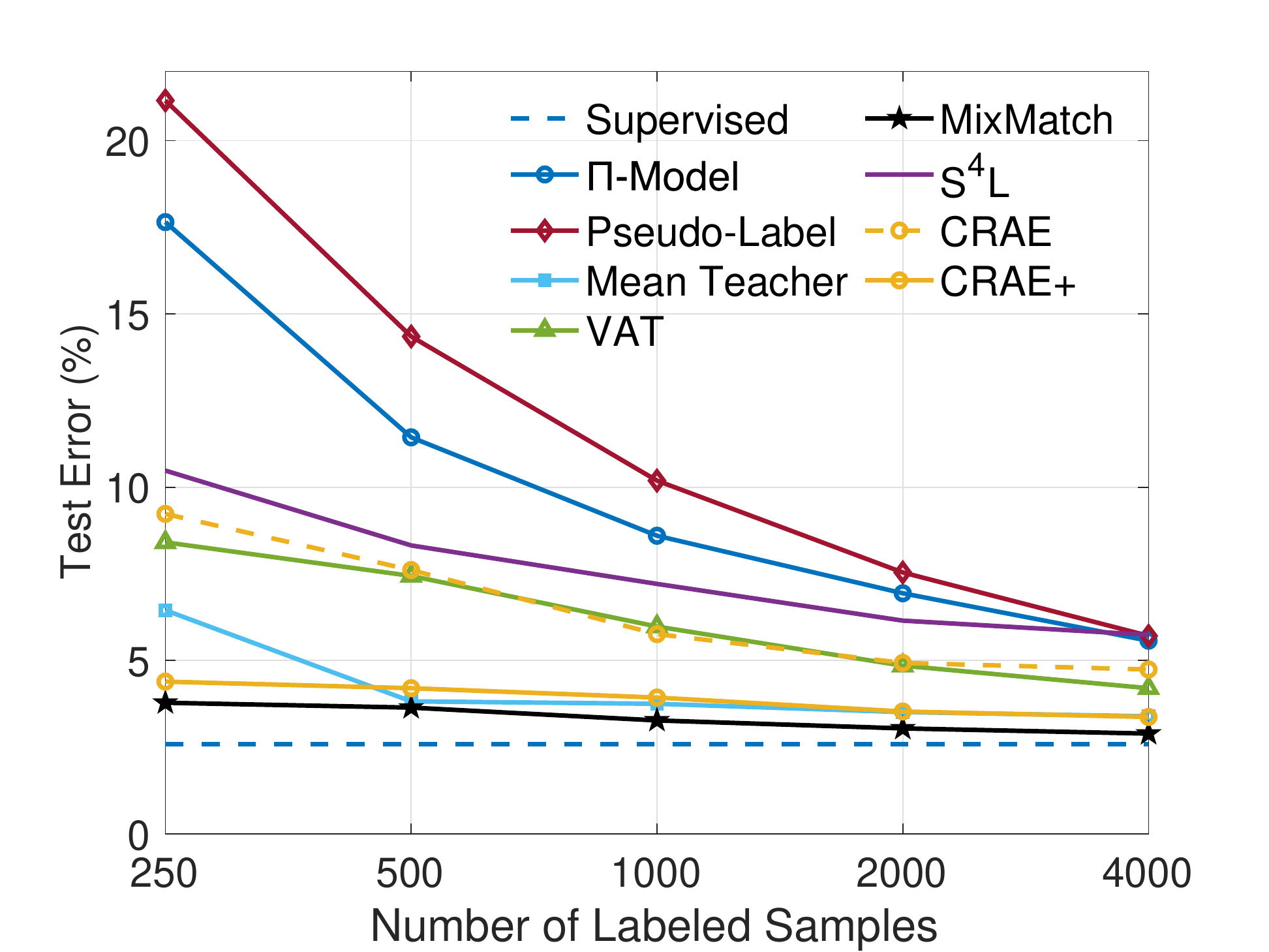}
\caption{SVHN error rates (\%) of CRAE+
and SOTA SemSL.}
\label{fig:svhn}
\end{figure}

The experimental results are shown in Figure~\ref{fig:cifar10}, Figure~\ref{fig:svhn} and Table~\ref{tab:cifar100_ssl}. As seen from those Figures and Table, by incorporating the proposed two extensions, CRAE+ achieves significant improvement over the basic CRAE. Moreover, we find that the proposed CRAE+ is on-par with the best performed approaches, e.g., Mixmatch, in those datasets. This clearly demonstrates the power of the proposed method. Note that the current state-of-the-art in SemSL is achieved by carefully combining multiple existing successful ideas in SemSL. In contrast, our CRAE+ achieves excellent results via an innovative framework of marrying SlfSL with SemSL. Conceptually, the latter enjoys greater potential. In fact, CRAE might be further extended by using more successful techniques in SemSL, such as mixup~\cite{verma2019interpolation}. Since the focus of this paper is to study how SlfSL can benefit SemSL, we do not pursue this direction. 

\subsection{Ablation Study}\label{exp:ablation_study}
Both CRAE and CRAE+ methods consist of multiple components. In this section, we conduct ablation studies to examine their impact and investigate how our methods work. Specifically, we first examine the importance of some key components in CRAE and CRAE+. Followed by an investigation on the relationship between CRAE and ensemble methods. Finally, we also study the effect of auxiliary semantic classifier to both CRAE and CRAE+ mehtods.

\noindent \textbf{1. The impact of various components in CRAE and CRAE+.}

We study these effects through adding or removing some components in order to provide additional insight into the role of each part in CRAE and CRAE+. Specifically, we measure the effect of (1) only adding extension 1 to CRAE, i.e., make the prediction of semantic classifier more sharpen.
(2) further adding extension 2 to CRAE. The resulted model is identical to CRAE+. (3) removing whole rotation angle prediction branches from CRAE, i.e., pure supervised method with data rotated. (4) removing rotation angle prediction branches and adding extension 1 to CRAE. The resulted structure can be seen as a variant of only using the SemSL technique in Extension 1.
\begin{table}[t]
\centering
\caption{Ablation study to the impact of two extensions for CRAE on CIFAR-10 with 250 and 4000 labels.}
\label{tab:ablation_study}
\begin{tabular}{l r r }
\hline
\hline
\# Labels & \multicolumn{1}{c}{250} & \multicolumn{1}{c}{4000} \\
\hline
CRAE & 17.07 & 9.07 \\

CRAE + ext. 1 & 12.68 & 7.05 \\

CRAE + ext. 1 + ext. 2 (=CRAE+)  & 10.76 & 5.98 \\
\hline
CRAE w/o rotation branches & 62.73 & 27.31 \\

CRAE w/o rotation branches + ext. 1 & 54.09 & 14.38 \\

\hline
\hline
\end{tabular}
\end{table}
\begin{figure}[t]
\centering
\includegraphics[width=\textwidth]{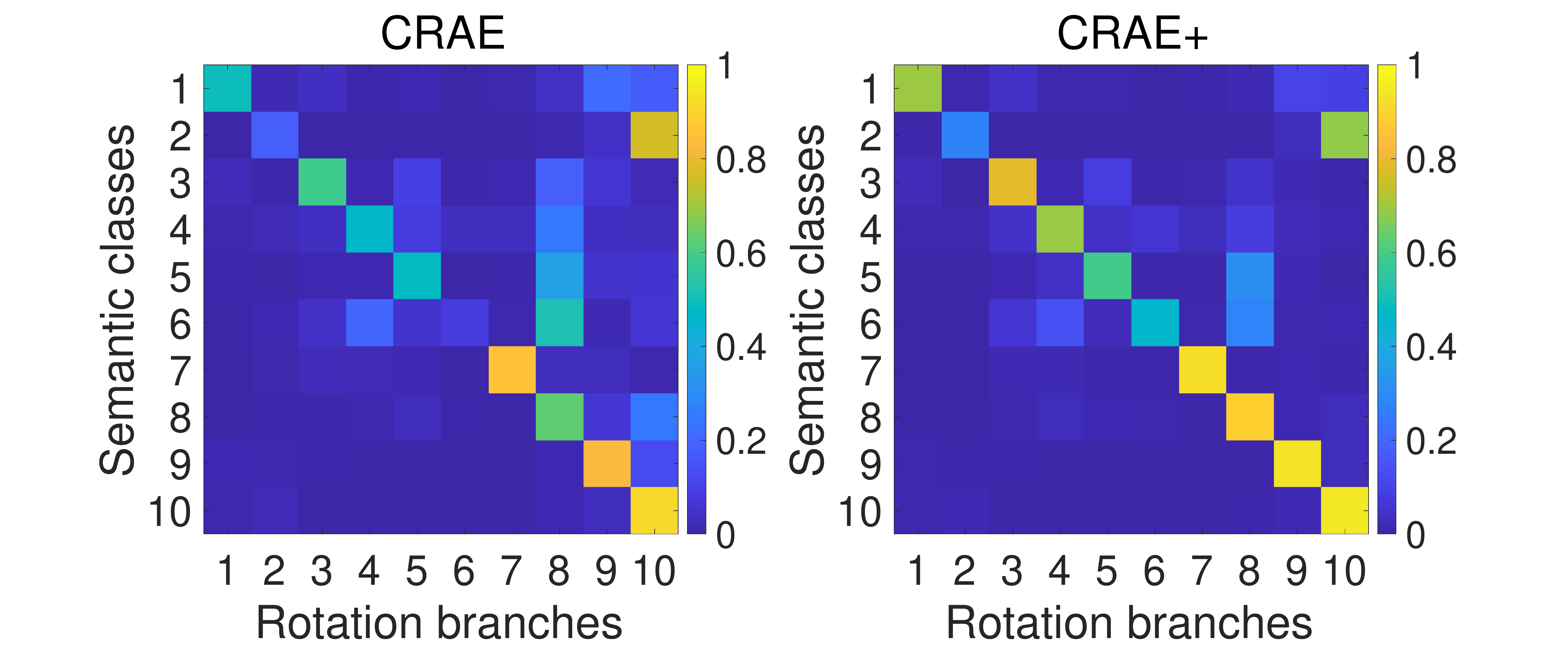}
\caption{Comparison of rotation predictors' diversity (based on the model at 1024 epoch) between CRAE and CRAE+ methods on CIFAR-10 with 250 labels.}
\label{fig:diversity_crae_vs_craeP}
\end{figure}

We conduct our studies on CIFAR-10 with 250 and 4000 labels with results presented in Table~\ref{tab:ablation_study}. The main observations are: (1) The two extensions in CRAE+ will bring varying degrees of improvement. In Figure~\ref{fig:diversity_crae_vs_craeP}, we also compare the confusion matrices (please see Figure~\ref{fig:diversity} and section~\ref{sect:analysis} for its definition) of CRAE and CRAE+. As seen, the confusion matrix of CRAE+ is closer to diagonal matrix, and it actually mitigates the confusion of some classes. This validates that CRAE+ indeed makes the conditional rotation estimator more specialized to its assigned class as expected.  
(2) Applying rotation as a data augmentation for labeled data will not lead to improved performance over the labeled-only baseline as by cross referring the results in Table~\ref{tab:cifar10}.
This shows that the advantage of CRAE is not coming from the rotation data augmentation.
(3) Using the sharpening strategy as in our extension 1 and training a SemSL method alone (the last method in Table~\ref{tab:ablation_study}) does not produce good performance.
This indicates the superior performance of CRAE+ is not simply coming from a strong SemSL method but its incorporation with the CRAE framework.

\noindent \textbf{2. The relationship to ensemble method.}

The structure of CRAE enjoys the benefit of ensembling rotation predictors. In order to investigate to what extent ensemble contributes to the good performance of CRAE, we compare CRAE with three ensemble baselines. The first one replaces the guidance from $p(y|x)$ with a multinomial random vector which means the rotation predictor is randomly selected. The second one independently trains all rotation estimation branches. The last one uses $p(y|x)$ to perform softly selection as in CRAE, but not back-propagates gradient to it. This operation is usually called ``detech''.  

From the results shown in Table~\ref{tab:ablation_ensemble}, we can see that using ensemble strategy indeed leads to improved performance compared to $S^4L$ method (numbers can be refered in Table~\ref{tab:cifar10}). Among three ensemble methods, using independent branches and detach $p(y|x)$ achieves the best performance. However, CRAE still achieves superior performance over them, especially when the number of labeled samples is small. This clearly shows that CRAE can be benefited from the indirect supervision on $p(y|x)$ generated during optimizing $p(z|x)$ and the good performance of CRAE cannot be simply attributed to the ensemble strategy. Moreover, with the proposed extensions, CRAE+ achieves significantly better performance than the ensemble baselines.

\begin{table}[t]
\centering
\caption{Ablation study to the relationship between CRAE and  ensemble methods on CIFAR-10 with 250 and 4000 labels.}
\label{tab:ablation_ensemble}
\begin{tabular}{l c c}
\hline
\hline
\# Labels & 250 & 4000\\
\hline
CRAE & 17.07 & 9.07 \\
Ensemble (random select) & 28.68 & 14.42 \\
Ensemble (independent) & 23.26 & 11.66 \\
CRAE detach $p(y|x)$ & 23.27 & 11.88\\
\hline
CRAE+ & 10.76 & 5.98\\
\hline
\hline
\end{tabular}
\end{table}
\begin{table}[t]
\centering
\caption{Ablation study to the effect of additional semantic classifier to CRAE and CRAE+ methods.}
\label{tab:ablation_study_aux_cls}
\begin{tabular}{l r r | r r}
\hline
\hline
& \multicolumn{2}{c}{CRAE} & \multicolumn{2}{c}{CRAE+} \\
use of add. cls. & \multicolumn{1}{c}{\cmark} & \multicolumn{1}{c |}{\xmark} & \multicolumn{1}{c}{\cmark} & \multicolumn{1}{c}{\xmark} \\
\hline
CIFAR10-250 & 17.07 & 18.33 & 10.76 & 11.37  \\
\hline
CIFAR10-4000 & 9.07 & 10.25 & 5.98 & 6.28 \\
\hline
\hline
\end{tabular}
\end{table}

\noindent \textbf{3. The benefit of using additional semantic classifier.}

As discussed in Section~\ref{sec:obstacles}, adding an additional semantic classifier branch to CRAE can slightly improve the performance. In this subsection, we conduct an ablation study to examine its impact on both CRAE and CRAE+. The results are shown in Table~\ref{tab:ablation_study_aux_cls}. As seen, for CRAE, the additional classifier will bring more than one percent improvement on both 250 and 4000 labels of CIFAR-10 dataset. But for CRAE+, the additional classifier brings marginal improvement. This is perhaps due to that the two extensions makes the supervision for $p(y|x)$ less noisy.

\section{Conclusion}
In this work, we propose an idea to couple SemSL with SlfSL. Implementing this idea with rotation-angle-estimation-based SlfSL, we design a new semi-supervised learning featured by conditional rotation angle estimation. Two extensions are further developed to enhance its performance. We show that the proposed method can achieve comparable performance to the state-of-the-art SemSL methods.

\section{Appendix}
\subsection{Algorithm details of CRAE and CRAE+}\label{app:alg_details}
The pseudocodes of CRAE and CRAE+ are presented in Algorithm~\ref{alg:crap} and Algorithm~\ref{alg:crap_plus} respectively.

\subsubsection{Experimental details}
The experimental details are presented in Table~\ref{tab:train_parameter}.

\subsubsection{Tabular results}
Table~\ref{tab:cifar10_ssl} and~\ref{tab:svhn_ssl} presents a summary for error rate comparison of CRAE and CRAE+ to existing SemSL methods on CIFAR-10 and SVHN respectively. Results in top block are reported in literature where mark $\dagger$ means that the results come from~\cite{oliver2018realistic}, mark $\ddagger$ means that the results come from~\cite{verma2019interpolation} and others are from~\cite{berthelot2019mixmatch}. Results locating in the bottom block are achieved by our implementation.
\begin{table}[h]%
\centering
\caption{Error rate (\%) comparison of CRAE
to existing SemSL methods on CIFAR-10.}
\label{tab:cifar10_ssl}
\begin{tabular}{ l r r r r r r }
\hline
\hline
\# Labels & \multicolumn{1}{c}{250} & \multicolumn{1}{c}{500} & \multicolumn{1}{c}{1000} & \multicolumn{1}{c}{2000} & \multicolumn{1}{c}{4000} \\
\hline
$\text{Labeled-only}^{\dagger}$ & - & - & - & - & 20.26 \\
$\Pi$-Model & 53.02 & 41.82 & 31.53 & 23.07 & 17.41 \\
Pseudo-Label & 49.98 & 40.55 & 30.91 & 21.96 & 16.21 \\
Mixup & 47.43 & 36.17 & 25.72 & 18.14 & 13.15 \\
VAT & 36.03 & 26.11 & 18.68 & 14.40 & 11.05 \\
MeanTeacher & 47.32 & 42.01 & 17.32 & 12.17 & 10.36 \\
MixMatch & 11.08 & 9.65 & 7.75 & 7.03 & 6.24 \\
$\text{ICT}^{\ddagger}$ & - & - & 15.48 & 9.26 & 7.29 \\
\hline
\hline
Labeled-only &  56.76 &	47.24 & 36.09 & 29.90 & 19.65\\

Fine-tune & 53.66 & 35.18 & 28.17 & 22.00 & 17.39 \\

$S^4L$ & 32.66 & 28.23 & 22.55 & 18.91 & 15.71 \\

CRAE & 17.07 & 15.57 & 12.90 & 10.72 & 9.07 \\

CRAE+ & 10.76 & 9.20 & 7.43 & 6.98 & 5.98\\
\hline
\hline
\end{tabular}
\end{table}
\begin{table}[h]%
\centering
\caption{Error rate (\%) comparison of CRAE
to existing SemSL methods on SVHN.}
\label{tab:svhn_ssl}
\begin{tabular}{ l r r r r r r }
\hline
\hline
\# Labels & \multicolumn{1}{c}{250} & \multicolumn{1}{c}{500} & \multicolumn{1}{c}{1000} & \multicolumn{1}{c}{2000} & \multicolumn{1}{c}{4000} \\
\hline
$\text{Labeled-only}^{\dagger}$ & - & - & 12.83 & - & - \\
$\Pi$-Model & 17.65 & 11.44 & 8.60 & 6.94 & 5.57 \\
Pseudo-Label & 21.16 & 14.35 & 10.19 & 7.54 & 5.71 \\
Mixup & 39.97 & 29.62 & 16.79 & 10.47 & 7.96 \\
VAT & 8.41 & 7.44 & 5.98 & 4.85 & 4.20 \\
MeanTeacher & 6.4 & 3.82 & 3.75 & 3.51 & 3.39 \\
MixMatch & 3.78 & 3.64 & 3.27 & 3.04 & 2.89 \\
$\text{ICT}^{\ddagger}$ & 4.78 & 4.23 & 3.89 & - & - \\
\hline
\hline
Labeled-only & 23.27 & 17.85 & 12.84 & 9.65 & 7.26 \\

Fine-tune & 16.21 & 11.54 & 8.31 & 7.30 & 5.61 \\

$S^4L$ & 10.48 & 8.32 & 7.21 & 6.15 & 5.74 \\

CRAE & 9.23 & 7.61 & 5.76 & 4.93 & 4.74 \\

CRAE+ & 4.10 & 4.21 & 3.91 & 3.59 & 3.36 \\
\hline
\hline
\end{tabular}
\end{table}

\subsection{Ablation study to the independent ensemble method}
We further conduct an ablation study to the independent ensemble method to test its performance on various numbers of rotation classifiers. According to the results shown in Table~\ref{tab:ablation_ens_head_num}, more rotation angle classifiers are limited to the promotion of the independent ensemble method.
\begin{table}[h]
\centering
\caption{Ablation study to the effect of rotation classifiers' number on independent ensemble method.}
\label{tab:ablation_ens_head_num}
\begin{tabular}{l c c c}
\hline
\hline
rot. cls. num. & 2 & 10 & 20 \\
\hline
CIFAR10-250 & 23.29 & 23.26 & 23.04\\
\hline
\hline
\end{tabular}
\end{table}
\begin{algorithm*}[h]
\caption{CRAE Pseudocode}
\label{alg:crap}
\begin{algorithmic}[1]
\REQUIRE ~~\\
$\{X_l, Y_l\}$: collection of labeled samples\\
$X_u$: collection of unlabeled samples\\
$f_\text{sem\_cls}$: semantic classifier\\
$f^{k}_\text{rot\_cls}$: $k$-th rotation angle classifier, where $k\in [1,2,\cdots, C]$\\
$T$: total number of iterations\\
$B$: minibatch size\\
\ENSURE ~~\\
$f_\text{sem\_cls}$: semantic predictor for test set\\
\hspace{-15pt}{\bf Process:}\\
    \FOR{$t \gets 1$ to $T$}
        \STATE Sample $B$ examples from $\{X_l, Y_l\}$ and $X_u$ respectively
        \STATE Obtain \textbf{semantic class prediction} on semantic classifier: $p(y|x_l)=f_\text{sem\_cls}(x_l)$ and train with $\text{CrossEntropy}(p(y|x_l), y_l)$ loss
        \STATE Obtain \textbf{rotation angle prediction} on labeled samples $p(z|x_l)=\sum_{y_l} p(z|y_l, x_l)\cdot \text{one\_hot}(y_l) = f^{y_l}_\text{rot\_cls}(x_l)$ and train with $\text{CrossEntropy}(p(z|x_l), \hat{z})$ loss \tcp{$\hat{z}$ is rotation angle ground truth}
        \STATE Obtain \textbf{rotation angle prediction} on unlabeled samples $p(z|x_u)=\sum_{y} p(z|y, x_u)\cdot p(y|x_u) = \sum^{C}_{k=1} f^{k}_\text{rot\_cls}(x_u)\cdot f^{k}_\text{sem\_cls}(x_u)$ and train with $\text{CrossEntropy}(p(z|x_u), \hat{z})$ loss %
    \ENDFOR
\end{algorithmic}
\end{algorithm*}
\begin{algorithm*}[h]
\caption{CRAE+ Pseudocode}
\label{alg:crap_plus}
\begin{algorithmic}[1]
\REQUIRE ~~\\
$\{X_l, Y_l\}$: collection of labeled samples\\
$X_u$: collection of unlabeled samples\\
$f_\text{sem\_cls}$: semantic classifier\\
$f^{k}_\text{rot\_cls}$: $k$-th rotation angle classifier, where $k\in [1,2,\cdots, C]$\\
$T$: total number of iterations\\
$B$: minibatch size\\
\ENSURE ~~\\
$f_\text{sem\_cls}$: semantic predictor for test set\\
\hspace{-15pt}{\bf Process:}\\
    \FOR{$t \gets 1$ to $T$}
        \STATE Sample $B$ examples from $\{X_l, Y_l\}$ and $X_u$ respectively
        \STATE Obtain \textbf{semantic class prediction} on semantic classifier: $p(y|x_l)=f_\text{sem\_cls}(x_l)$ and train with $\text{CrossEntropy}(p(y|x_l), y_l)$ loss 
        \STATE Construct sharpened target $\hat{p}$ for unlabeled data \tcp{see extension 1 in Section~3.4 of main text}
        \STATE Obtain \textbf{semantic class prediction for unlabeled data} on semantic classifier $p(y_u|x_u) = f_\text{sem\_cls}(x_u)$ and train with $\text{CrossEntropy}(p(y_u|x_u), \hat{p})$ loss
        \STATE Construct mixed images $\{x^m_l, x^m_u\}$ from $\{X_l, X_u\}$ \tcp{see extension 2 in Section~3.4 of main text}
        \STATE Obtain \textbf{rotation angle prediction} on labeled samples $p(z_m|x^m_l)=\sum_{y_l} p(z_m|y_l, x^m_l)\cdot \text{one\_hot}(y_l) = f^{y_l}_\text{rot\_cls}(x^m_l)$ and train with $\text{CrossEntropy}(p(z_m|x^m_l), \hat{z})$ loss \tcp{$\hat{z}$ is rotation angle ground truth}
        \STATE Obtain \textbf{rotation angle prediction} on unlabeled samles $p(z_m|x^m_u)=\sum_{y} p(z_m|y, x^m_u)\cdot p(y|x_u) = \sum^{C}_{k=1} f^{k}_\text{rot\_cls}(x^m_u)\cdot f^k_\text{sem\_cls}(x_u)$ and train with $\text{CrossEntropy}(p(z_m|x^m_u), \hat{z})$ loss
        
    \ENDFOR
\end{algorithmic}
\end{algorithm*}
\begin{table*}[h]%
\centering
\caption{Experimental details.}
\label{tab:train_parameter}
\begin{tabular}{l c c c c c }
\hline
\hline
datasets & \multicolumn{1}{c}{CIFAR-10} & \multicolumn{1}{c}{SVHN/+Extra} & \multicolumn{1}{c}{CIFAR-100} & \multicolumn{1}{c}{ILSVRC2012-1\%} & \multicolumn{1}{c}{ILSVRC2012-10\%} \\
\hline
architecture & WRN-28-2 & WRN-28-2 & WRN-28-8.4375 & ResNet50v2 & ResNet50v2 \\
\# training set & 50000 & 73257/+531131 & 50000 & 1281167 & 1281167 \\
\# labeled set & \multicolumn{2}{c}{\{250, 500, 1000, 2000, 4000\}} & 10000 & 13762 & 128866 \\
\# validation set & 5000 & 7325 & 5000 & 5005 & 50046\\
minibatch size & 64 & 64 & 64 & 256 & 256 \\
optimizer & Adam & Adam & Adam & SGD & SGD \\
LR & 0.002 & 0.002 & 0.002 & 0.01 & 0.1\\
weight decay & 0.02 & 0.02/0.0001 & 0.04 & 0.01 & 0.001 \\
\# epoch & 1024 & 500 & 300 & 1000 & 200 \\
\# iteration/epoch & 1024 & 1024 & 1024 & 53 & 503 \\
LR rampup & \xmark & \xmark & \xmark & 10 epoch & 5 epoch \\
LR decay & \xmark & \xmark & \xmark & 10 & 10 \\
LR decay at & \xmark & \xmark & \xmark & \{700,800,900\} & \{140,160,180\} \\
EMA model & \cmark & \cmark & \cmark & \xmark & \xmark \\
\hline
\hline
\end{tabular}
\end{table*}

{\small
\bibliographystyle{ieee_fullname}
\bibliography{egbib}
}

\end{document}